%% file: main_document.tex
\documentclass[runningheads]{llncs}

\usepackage[numbers]{natbib}
\usepackage{graphicx}
\usepackage{float}
\usepackage{amsmath,amssymb,mathrsfs}
\usepackage[dvipsnames,svgnames]{xcolor}
\usepackage{setspace}
\usepackage{array,adjustbox}
\usepackage{xurl}
\usepackage[hidelinks]{hyperref}
\begin{document}

\title {An ensemble of graph generative models for assisted history matching of the Brugge field case}

\author{Gleb Shishaev\inst{1} \and
Anton Shirokov\inst{1} \and
Shadfar Davoodi\inst{1,2} \and
Mohammed Al-Shargabi\inst{1} \and
Vasily Demyanov\inst{3} \and
Daniel Arnold\inst{3}}

\titlerunning{Graph generative ensemble for AHM on Brugge}
\authorrunning{G. Shishaev et al.}

\institute{
Tomsk Polytechnic University, Tomsk, Russia \and
Skolkovo Institute of Science and Technology, Moscow, Russia \and
Heriot-Watt University, Edinburgh, United Kingdom
}

\maketitle 

\input{abstract}
 
\input{chapters/Introduction}
\input{chapters/Brugge_case}

\input{chapters/Methodology}
\input{chapters/Application}
\input{chapters/Results}
\input{chapters/Conclusions}
\input{chapters/Acknowledgments}

\section*{Author contributions}
\noindent\textbf{Gleb Shishaev:} Conceptualisation, Methodology, Software, Formal analysis, Investigation, Data curation, Visualisation, Project administration, Writing - original draft, Writing - review \& editing.\par
\noindent\textbf{Anton Shirokov:} Methodology, Software, Formal analysis, Investigation, Validation, Visualisation, Writing - review \& editing.\par
\noindent\textbf{Shadfar Davoodi:} Formal analysis, Investigation, Validation, Writing - review \& editing.\par
\noindent\textbf{Mohammed Al-Shargabi:} Investigation, Validation, Writing - review \& editing.\par
\noindent\textbf{Vasily Demyanov:} Conceptualisation, Methodology, Supervision, Project administration, Writing - review \& editing.\par
\noindent\textbf{Daniel Arnold:} Conceptualisation, Methodology, Validation, Supervision, Writing - review \& editing.\par

\section*{Code availability}

The source codes are available for downloading at the link:
\url{https://github.com/GlebShish/GWAE_Brugge}

\section*{Data availability}

The prior ensemble of 10{,}000 geological realisations generated for this study is available from the corresponding author upon reasonable request.

\section*{Declaration of competing interest}

The authors declare the following financial interests/personal relationships which may be considered as potential competing interests: Vasily Demyanov reports that support and access to tNavigator reservoir simulator was provided by Rock Flow Dynamics. The remaining authors declare that they have no known competing financial interests or personal relationships that could have appeared to influence the work reported in this paper.

\bibliographystyle{unsrtnat}
\bibliography{bibliography}

\input{sup_materials}

\end{document}

%% file: abstract.tex
\begin{abstract}
Assisted history matching remains non-unique because many geological realisations can reproduce equivalent production responses with substantial differences between individual well matches. This ambiguity increases when a field comprises several depositional settings and modelling strategies. The paper applied a graph autoencoder approach to history matching of a benchmark Brugge field that comprises different depositional zones. Each stratigraphic zone is modelled with its dedicated generative model that comprises a joint latent space for optimisation history matching loops. Graph-based generators operate directly on the active cells of the corner-point grid, so its irregular geometry is handled without conversion to a regular lattice. The optimisation search in the 160-dimensional joint latent space is done with the Covariance Matrix Adaptation Evolution Strategy. The multi-component objective function includes static well data mismatch, dynamic production mismatch, and a geological realism penalty based on the \textit{Decoder} induced latent metric. Latent space diagnostics with topological data analysis are applied prior to history matching to test whether one generator is sufficient or whether a multiple generator formulation is required. The results show that the zone-based formulation gives coherent latent spaces, preserves the main porosity and permeability relations that vary from one zone to another, and supports history matching while keeping the solutions within the learned geological manifold. The matched models reproduce field oil and water production within the prior uncertainty range and provide a comparable well-level match to previously published Brugge studies based on Ensemble Kalman Filter, Particle Swarm Optimisation with principal component analysis of parameterisation, and Multiple Kernel Learning with Particle Swarm Optimisation.

\keywords{History matching \and Graph generative models \and Geological realism \and Heterogeneous geological prior}

\end{abstract}

%% file: chapters/Introduction.tex
\section{Introduction}
\label{sec:intro}

Assisted history matching (AHM) updates reservoir property fields so that simulated responses match observed field data. Because these observations constrain the reservoir indirectly and sparsely, different geological models can yield similar responses, and practical AHM must balance an acceptable data match, uncertainty representation, and geological consistency. The parameterisation must therefore permit dynamic calibration while remaining consistent with static data, geological structure, and petrophysical relationships, especially for field-scale corner-point grids \citep{aanonsen2009ensemble,canchumuni2021recent}.

All AHM workflows depend on a prior defined by geological interpretations and conditioning data. For generative parameterisations, the learned model cannot recover geological features absent from the training ensemble \citep{canchumuni2021recent}.

Different parameterisation strategies address this trade-off in different ways, as illustrated by Brugge benchmark studies using EnKF, PSO with PCA parameterisation, kernel methods, and ensemble smoothers \citep{peters2010results,mohamed2010application,demyanov2015mkl,cruz2022joint}. Geostatistical workflows honour well data and prescribed geological rules but may lack freedom for dynamic calibration, whereas ensemble methods update property fields through sample covariances that may break facies boundaries, property relationships, and channel connections. Regularisation and covariance localisation reduce unstable updates but do not define the full set of admissible geological models, so neither approach can guarantee geological consistency by itself \citep{aanonsen2009ensemble,canchumuni2021recent,canchumuni2019towards}.

Deep generative models offer an alternative parameterisation: an \textit{Encoder} maps each realisation to a compact latent vector, and a \textit{Decoder} reconstructs the property fields, without replacing the geomodelling methods used to construct the prior \citep{jo2022generative,canchumuni2019towards,difederico2025latent}. For AHM, the optimiser then updates this compact vector instead of thousands of grid-cell values, reducing the search dimension and linking updates to patterns learned from the prior.

Several generative architectures have been explored for subsurface parameterisation, including generative adversarial networks \citep{goodfellow2014,mosser2017reconstruction,laloy2018training}, variational autoencoders and ensemble-smoother extensions \citep{kingma2014,canchumuni2019towards,canchumuni2021recent,bao2022variational}, CNN-PCA \citep{liu2021cnnpca}, and latent diffusion models \citep{difederico2025latent}.

However, many such workflows assume regular lattice grids, whereas realistic reservoir models commonly use irregular corner-point grids with faults and non-uniform cell geometry \citep{Roe2016Volume}. Mapping these models to a lattice can introduce geometric artefacts and weaken the link to the simulator grid \citep{Chen2017CornerPoint}, motivating generative approaches that operate on the native grid.

In an earlier study by the authors, we introduced a Graph-based Wasserstein Autoencoder workflow for history matching, referred to here as GWAE, in which the geological model on its native grid is represented as a graph and the latent variables of the generative model define the parameter space for optimisation \citep{shishaev2025history}. In synthetic settings, GWAE showed that graph-based latent representations can support history matching while retaining geological structure.

The present study extends this formulation to a field-scale Brugge benchmark case with a composite prior, where both the reservoir geometry and the prior structure are more complex than in synthetic demonstrations \citep{peters2010results}. The study is focused on two aspects: (i) irregular corner-point unstructured grid that accounts for realistic geological structure and makes the graph modelling approach more relevant; (ii) a variety of depositional environments that represent a wide range of geological scenario variations.  

The second challenge concerns the composite structure of the Brugge prior: each realisation comprises four stratigraphic zones with distinct depositional settings and property distributions, and a separate diagnostic Schelde dataset combines Object Modelling (OM) and Sequential Indicator Simulation (SIS) realisations to test whether both strategies can share one latent representation. After training, the \textit{Encoder} maps each realisation to a latent vector, and the resulting latent point cloud reveals how the \textit{Encoder} organises these geological groups.

If the latent point cloud contains separated components, the paths between them may cross regions of low model density, where the \textit{Decoder} may be less reliable \citep{shishaev2025history}. The representativeness of the latent space training samples is evaluated with topological analysis to understand how well the latent space represents multiple geomodelling concepts. 


The main contributions of the Brugge study presented in this paper are summarised as follows:
\begin{itemize}
    \item \textbf{Graph-based AHM on the native Brugge grid}

    \item \textbf{A zone-wise ensemble of graph generators for each distinct Brugge sedimentary setting}

    \item \textbf{Interpretable topological diagnostics for latent parameterisation}

    \item \textbf{Geological-realism-aware calibration combines static mismatch, dynamic mismatch, and a \textit{Decoder}-based realism term}
\end{itemize}

%% file: chapters/Brugge_case.tex
\section{Brugge case study and dataset construction}
\label{sec:brugge_case}

\subsection{The Brugge benchmark}
\label{sec:brugge_benchmark}

A semi-synthetic Brugge benchmark case was used for the history matching application \citep{peters2010results}. Brugge was designed for  history matching and uncertainty quantification stidies and provides a realistic ensemble of hetregenous reservoir proprty across different depositional setting. The reservoir consists of four stratigraphic zones---Fluvial (Schelde), Lower shoreface (Maas), Upper shoreface (Waal), and Sandy shelf (Schie)---summarised in Table~\ref{tab:brugge_zones}. The field is operated under water injection using 30 vertical wells (20 producers, 10 injectors); Waal contributes approximately 80\% of total fluid production, while Schie contributes below 1\% (Figure~\ref{fig:hm_stats}(a)).

\begin{table}[htbp]
\caption{Main characteristics of the four Brugge stratigraphic zones, after \citet{peters2010results}.}
\label{tab:brugge_zones}
\centering
\begingroup
\small
\setstretch{1}
\setlength{\tabcolsep}{5pt}
\renewcommand{\arraystretch}{1.1}
\begin{adjustbox}{max width=\textwidth}
\begin{tabular}{lllll}
\hline
Zone & \shortstack{Avg. thickness,\\m} & \shortstack{Avg. porosity,\\\%} & \shortstack{Avg. permeability,\\mD} & \shortstack{Depositional\\environment} \\
\hline
Schelde & 10 & 21 & 1100 & Fluvial \\
Maas    & 20 & 19 & 90   & \shortstack{Lower\\shoreface} \\
Waal    & 26 & 24 & 814  & \shortstack{Upper\\shoreface} \\
Schie   & 5  & 19 & 36   & Sandy shelf \\
\hline
\end{tabular}
\end{adjustbox}

\endgroup
\end{table}

The Brugge model is defined on a $113 \times 38 \times 19$ corner-point grid containing 81,586 cells. Although the grid has regular $(i,j,k)$ indexing, fault offsets, stratigraphic thickness variations, and the non-rectangular active-cell mask produce non-uniform physical geometry and connectivity; 69.5\% of the cells are active. This makes Brugge directly relevant to the graph formulation, which operates on the native active-cell topology without lattice resampling.

\subsection{Prior dataset construction}
\label{sec:prior_dataset}

The public Brugge ensemble includes only 104 realisations, too few for training generative models. A zone-wise training dataset of 10,000 realisations was therefore generated following the geological modelling principles of \citet{peters2010results}. This exparnded ensemble defines the support of each zone's learned latent space (LS).

The Schelde zone represents a fluvial environment with high permeability channel bodies embedded in a lower permeability background. In the original Brugge ensemble, \citet{peters2010results} used two geological modelling algorithms for the Schelde zone: OM and SIS. The Maas, Waal, and Schie zones were modelled only with SIS, using zone-specific settings. The Schelde prior, therefore, combines two plausible descriptions of the same depositional environment: OM represents the fluvial system through explicit channel bodies, whereas SIS represents it through variogram-controlled facies continuity, making Schelde a natural diagnostic case for testing whether one latent representation can describe a mixed prior (see Section 4).

For all zones, porosity was generated using Sequential Gaussian Simulation. Permeability was then assigned using one of three scenarios: 1) deterministically from porosity, 2) deterministically by facies, and 3) stochastically by co-kriging on porosity. The channel-geometry and variogram ranges used to generate the zone-wise training datasets are summarised in Table~\ref{tab:prior_ranges}. The Schelde SIS settings refer to the additional realisations used only in the mixed-prior diagnostic study (Section 4.1).

\begin{table}[htbp]
\caption{Channel-geometry and variogram ranges used to generate training datasets.}
\label{tab:prior_ranges}
\centering
\begingroup
\footnotesize
\setstretch{1}
\setlength{\tabcolsep}{3pt}
\renewcommand{\arraystretch}{1.1}
\begin{adjustbox}{max width=\textwidth}
\begin{tabular}{@{}>{\raggedright\arraybackslash}p{0.18\textwidth}>{\raggedright\arraybackslash}p{0.42\textwidth}>{\raggedright\arraybackslash}p{0.36\textwidth}@{}}
\hline
Zone & Parameter & Range \\
\hline
Schelde (OM) & Channel width / thickness & \mbox{[50; 550]} / \mbox{[5; 20]} \\
              & Orientation / amplitude / wavelength & \mbox{[200; 300]} / \mbox{[200; 1600]} / \mbox{[250; 2500]} \\
              & Porosity variogram, major/minor/vertical & \mbox{[1200; 1700]} / \mbox{[600; 1200]} / \mbox{[1; 5]} \\
Schelde (SIS) & Channel facies variogram, major/minor/vertical & \mbox{[350; 550]} / \mbox{[50; 250]} / \mbox{[5; 20]} \\
               & Porosity variogram, major/minor/vertical & \mbox{[1200; 1700]} / \mbox{[600; 1200]} / \mbox{[1; 5]} \\
Maas (SIS)    & Sand facies variogram, major/minor/vertical & \mbox{[500; 1500]} / \mbox{[400; 1050]} / \mbox{[1; 6]} \\
               & Porosity variogram, major/minor/vertical & \mbox{[1500; 2600]} / \mbox{[500; 2000]} / \mbox{[1; 4]} \\
Waal (SIS)    & Sand facies variogram, major/minor/vertical & \mbox{[1000; 1700]} / \mbox{[700; 900]} / \mbox{[20; 45]} \\
               & Porosity variogram, major/minor/vertical & \mbox{[1000; 3000]} / \mbox{[500; 1500]} / \mbox{[1; 6]} \\
Schie (SIS)   & Sand facies variogram, major/minor/vertical & \mbox{[400; 1500]} / \mbox{[200; 800]} / \mbox{[1; 6]} \\
               & Porosity variogram, major/minor/vertical & \mbox{[600; 1400]} / \mbox{[600; 1200]} / \mbox{[1; 6]} \\
\hline
\end{tabular}
\end{adjustbox}

\noindent\footnotesize
\textit{Note:} Channel orientation is reported in degrees; all channel dimensions and variogram ranges are reported in metres.
\endgroup
\end{table}

%% file: chapters/Methodology.tex
\section{Methodology}
\label{sec:methodology}

AHM is formulated here as a search for reservoir property fields that are consistent with static well data, dynamic production data, and the geological prior. Conventional geostatistical and ensemble-based AHM approaches, such as EnKF and ensemble smoothers, update reservoir models by regenerating and repopulating porosity and permeability fields at every iteration, either through direct geostatistical simulation or through covariance-based ensemble updates. The proposed workflow instead searches a lower-dimensional latent space learned from a prior ensemble of geological realisations. This reduces the dimensionality of the inverse problem and restricts the search mainly to geological variability represented in the prior.

The latent space is learned by a Graph-based Wasserstein Autoencoder trained on porosity and permeability fields from the prior ensemble of realisations. After training, the \textit{Decoder} maps a latent vector to full porosity and permeability realisations on the reservoir grid. Each decoded model is then passed to the flow simulator, and the optimisation algorithm updates the latent vector by minimising an objective function that combines production mismatch, static well-data mismatch, and a geological realism term.

The methodology follows five steps:
\begin{itemize}
    \item \textbf{Graph construction}:
    The native reservoir grid is converted into a graph representation used by the GWAE, preserving the active-cell topology and face-contact connectivity of the corner-point grid.
    \item \textbf{Latent space construction}:
    Graph representations are encoded by the GWAE into a continuous latent space, with each realisation of the prior mapped to a compact latent vector.
    \item \textbf{Latent space diagnostics}:
    The structure of the trained latent space is analysed using PCA, t-SNE, and topological data analysis to decide whether the prior should be represented by one generator or decomposed into several generators.
    \item \textbf{Multi-generator and hybrid latent space construction}:
    Where decomposition is warranted, separate generators are trained for coherent components of the prior, and their latent vectors are concatenated into a joint hybrid latent space.
    \item \textbf{History matching}:
    The hybrid latent parameterisation is optimised against a combined objective function integrating static mismatch, dynamic mismatch, and geological realism.
\end{itemize}

\subsection{Graph-based Wasserstein autoencoder on the native reservoir grid}
\label{sec:gwae_grid}

A defining feature of the proposed workflow is that the generative model operates directly on the native reservoir grid rather than on a resampled rectangular array. As discussed in Section~\ref{sec:intro}, field-scale reservoir models commonly use irregular corner-point grids, and resampling them to a regular lattice can distort cell geometry and weaken the link between the generated property fields and the simulator grid where flow physics is solved. To preserve this link, each reservoir realisation is first converted into an attributed graph by a fixed mapping:

\begin{equation}
\Phi : \mathcal{M} \rightarrow \mathcal{G}, \qquad \mathcal{G} = (V,E,\Psi) 
\end{equation}
where $\mathcal{M}$ denotes the native corner-point reservoir model: the grid geometry, the active-cell mask, and the porosity and permeability values assigned to the cells. The graph $\mathcal{G}$ is the representation used by the neural network. The nodes $V$ correspond to active reservoir cells, the edges $E$ describe physical face contact between active cells, and the node feature map $\Psi$ stores the cell properties used in this study, namely porosity and permeability. 

The mapping $\Phi$ is applied to all prior realisations before training. After training, the \textit{Decoder} generates porosity and permeability as node features on the same fixed active-cell graph, which are written back to the corresponding cells for flow simulation. Since grid geometry and the active-cell mask are fixed, the model parameterises only porosity and permeability, not the grid structure, and inactive cells are simply removed from the graph rather than retained as padding or masked nodes.

Graph edges are created from geometric cell adjacency, not from index adjacency in the structured grid array. Two active cells are connected by an edge whenever their opposing faces share an intersection with positive polygonal area, regardless of whether the cells are adjacent in index space. This definition is suitable for corner-point grids, where faults, local distortions, non-uniform cell geometry, and partial face overlaps can make physical connectivity differ from regular lattice connectivity.

This is the main practical difference from a lattice-based generator such as a convolutional autoencoder, which requires a rectangular array and must carry inactive cells as padding or masking (about 30\% of the Brugge array), while still failing to represent the true face-contact topology of a faulted corner-point grid. The graph representation avoids these limitations by storing only active cells and convolving over their actual connectivity.

The GWAE \textit{Encoder} and \textit{Decoder} are built from graph convolution operations defined on this active cell graph. For a target node $t$, the layer update aggregates information from its neighbours through trainable matrices \citep{kipf2017,morris2021},
\begin{equation}
h^{(l+1)}_t =
\sigma\!\left(
W^{(l)} h^{(l)}_t + Q^{(l)} \bar{h}^{(l)}_{\mathcal{N}(t)}
\right),
\label{eq:graph_update}
\end{equation}
where $h^{(l)}_t$ is the node feature vector at layer $l$, $\bar{h}^{(l)}_{\mathcal{N}(t)}$ is the aggregated feature vector over the neighbourhood $\mathcal{N}(t)$ defined by the graph edges, $W^{(l)}$ and $Q^{(l)}$ are trainable matrices, and $\sigma$ is a nonlinear activation. Because $\mathcal{N}(t)$ is determined by the face contact topology of the native simulation grid, the same convolution principle applies to all stratigraphic zones of Brugge without resampling. The \textit{Encoder} and \textit{Decoder} stacks follow \citet{shishaev2025history}; Table~\ref{tab:compression_dimensions} summarises the dimensions.

\begin{table}[htbp]
\caption{Grid, active-node feature and latent dimensions for the four Brugge zones.}
\label{tab:compression_dimensions}
\centering
\begingroup
\small
\setstretch{1}
\setlength{\tabcolsep}{5pt}
\renewcommand{\arraystretch}{1.1}
\begin{adjustbox}{max width=\textwidth}
\begin{tabular}{lrrrr}
\hline
Zone & \shortstack{Grid\\dimensions} & \shortstack{Grid\\cells} & \shortstack{Graph-node\\features} & Latent \\
\hline
Schelde & $113\times38\times5$  & 21,470 & $14{,}925\times2$ & 30 \\
Maas    & $113\times38\times5$  & 21,470 & $14{,}925\times2$ & 50 \\
Waal    & $113\times38\times7$  & 30,058 & $20{,}895\times2$ & 30 \\
Schie   & $113\times38\times2$  & 8,588  & $5{,}970\times2$  & 50 \\
\shortstack[l]{Full field /\\hybrid} & $113\times38\times19$ & 81,586 & $56{,}715\times2$ & 160 \\
\hline
\end{tabular}
\end{adjustbox}

\footnotesize\noindent
Both properties are jointly encoded into one latent vector per zone. Graph dimensions denote the node-feature matrix with two columns (porosity and permeability); grid geometry and connectivity are fixed.
\endgroup
\end{table}

The Wasserstein autoencoder objective combines reconstruction quality with regularisation of the latent distribution \citep{tolstikhin2018}:
\begin{equation}
D_{WAE}(P_X, P_G) =
\inf
\left[
\lambda D(Q_Z, P_Z)
+
\mathbb{E}_{P_X}
\mathbb{E}_{Q(Z|X)}
c(X, G(Z))
\right],
\label{eq:wae_objective}
\end{equation}
where $Q(Z|X)$ is the \textit{Encoder} distribution, $P_Z$ the latent prior, $G(Z)$ the \textit{Decoder}, $c$ the reconstruction cost, and $\lambda$ the regularisation strength. The latent prior is a standard Gaussian, and the regularisation term is implemented with Maximum Mean Discrepancy \citep{gretton2012} using an inverse multiquadratic kernel. The Wasserstein formulation is used because it encourages a smoother latent space than a standard variational autoencoder, which is important for history matching, where the optimiser must move through latent space and generate realistic intermediate realisations.

\subsection{Latent-space diagnostics for generator decomposition}
\label{sec:diagnostics}

The structure of a trained LS reflects how the prior ensemble is organised. Similar geological realisations are typically encoded close together, whereas distinct groups or scenarios may appear as separated regions of the latent point cloud. The connectedness of the encoded latent point cloud indicates whether the prior ensemble can be treated as a single coherent population or whether it comprises several distinct components.

This diagnostic is applied before history matching to select the parameterisation: whether a reservoir with multiple depositional zones should be represented by a single generator trained on the combined prior or decomposed into zone-specific generators with independent latent spaces updated jointly, with decoded zone fields stacked before flow simulation. This matters because latent-space optimisation assumes continuous movement through training-supported regions. A single encoder spanning distinct populations risks producing a latent space with disconnected, poorly supported regions, which the diagnostics below are designed to detect.

To determine whether the prior should be represented by a single generator or decomposed into separate generators, the encoded latent point cloud is analysed using complementary geometric and topological diagnostics. PCA \citep{jolliffe2016pca} provides a linear projection of the latent vectors, and t-distributed stochastic neighbour embedding (t-SNE) \citep{vandermaaten2008tsne} provides a nonlinear visualisation of local neighbourhood structure. Persistent homology is then used to quantify connectedness of the latent point cloud \citep{edelsbrunner2002, carlsson2009topology}. The persistence diagram of the zeroth homology group, $H_0$, provides the primary diagnostic: it records the filtration scales at which initially separate connected components merge \citep{shishaev2025history}.

Although persistence is computed in different homological dimensions, only $H_0$ features are used to select the generator parameterisation. For the latent point clouds considered here, higher homological dimensions have low persistence and are not used in the selection criterion.

The generator parameterisation is selected from the connectedness of the encoded latent point cloud. If a GWAE trained on the relevant prior yields a single connected and sufficiently dense representation, the single-generator formulation is retained. Conversely, persistent separated $H_0$ components indicate that the prior should not be treated as one coherent population but decomposed into more coherent components, each represented by a separate generator.

\subsection{Multiple generator formulation and hybrid latent space}
\label{sec:multigenerator}

For the Brugge case, a separate GWAE was trained for each stratigraphic zone of the prior ensemble. The latent dimensions in Table~\ref{tab:compression_dimensions} were selected by progressively reducing them until validation reconstructions showed increased error and loss of spatial or petrophysical features \citep{shishaev2025history}.
The hybrid latent vector is formed by concatenating the four zone latent vectors:

\begin{equation}
\mathbf{z}_{\mathrm{hybrid}} =
[\mathbf{z}_{\mathrm{Schelde}},
 \mathbf{z}_{\mathrm{Maas}},
 \mathbf{z}_{\mathrm{Waal}},
 \mathbf{z}_{\mathrm{Schie}}]
\in \mathbb{R}^{160}.
\end{equation}

\begin{figure}[htbp]
\centering
\includegraphics[width=0.90\linewidth]{"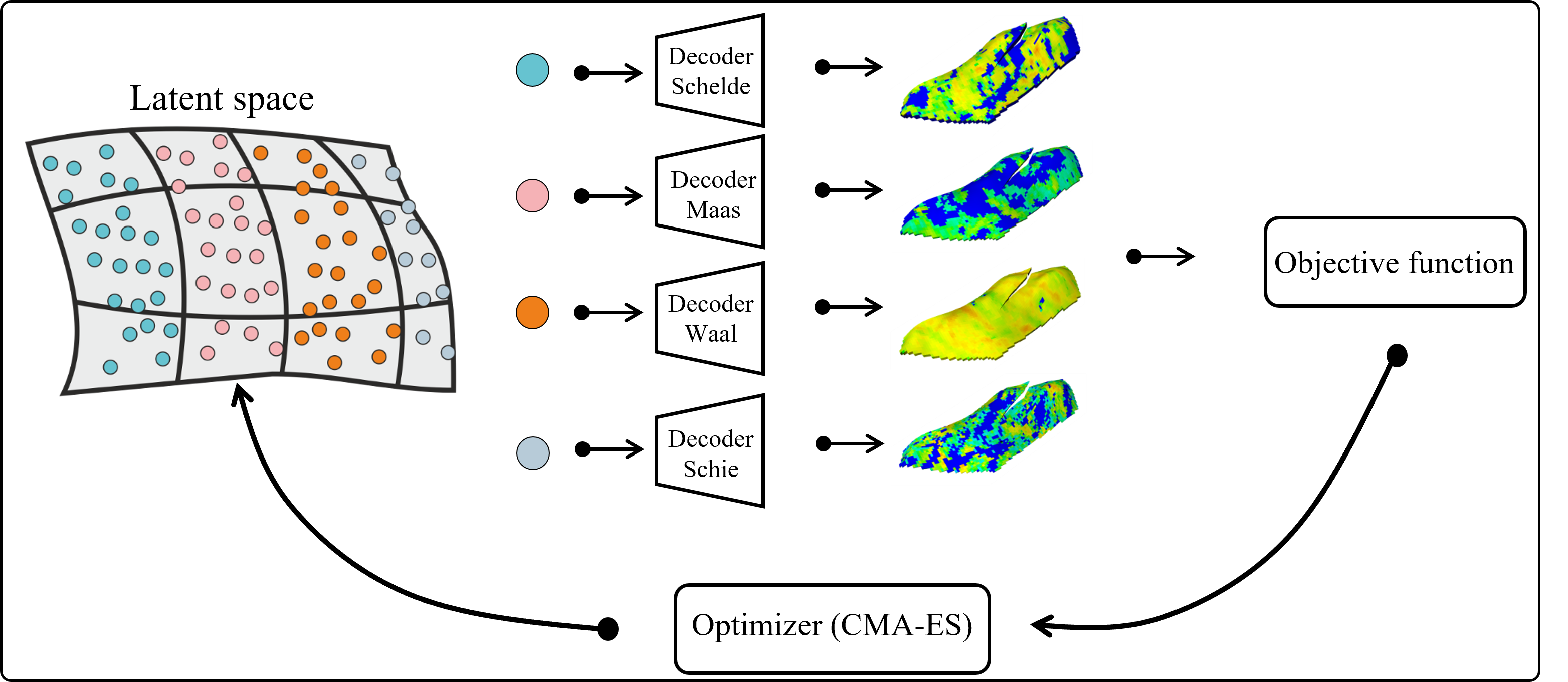"}
\caption{General scheme of inverse modelling in the hybrid latent space. The latent vector is divided into four parts that correspond to the Schelde, Maas, Waal, and Schie zones. Each part is decoded by the corresponding GWAE \textit{Decoder}. The generated zone models are stacked into a full reservoir realisation and evaluated by the objective function. The optimiser then proposes the next latent vector. Adapted from \citet{shishaev2024thesis}.}
\label{fig:inverse_model_scheme}
\end{figure}

During history matching, the optimiser searches directly in this joint latent space. At each evaluation, the hybrid vector is divided into four parts, each part is passed through the \textit{Decoder} trained for the corresponding zone, and the decoded zone property fields are stacked in stratigraphic order to form a complete Brugge reservoir model (Figure~\ref{fig:inverse_model_scheme}). This formulation assumes that the property distribution in one zone does not directly depend on the realisation in another zone.

This is consistent with the construction of the Brugge prior, in which zones were generated separately without cross-zone coupling. If strong dependencies must be represented, the framework can be extended by coupling the zone generators.

\subsection{Optimisation procedure and objective function}
\label{sec:optimisation}

History matching in the hybrid latent space is performed using the Covariance Matrix Adaptation Evolution Strategy (CMA-ES) \citep{hansen2001}. CMA-ES is suitable because it is derivative-free, which is essential when the objective function includes a non-differentiable flow-simulation step. Its multivariate Gaussian search distribution is also consistent with the standard Gaussian latent prior $P_Z$ and the Wasserstein regularisation that encourages the aggregated encoder distribution $Q_Z$ to match this prior (Section~\ref{sec:gwae_grid}). 

The Gaussian distribution is used only to propose candidate latent vectors: CMA-ES adapts its covariance matrix from the ranked candidates at each iteration, allowing the search distribution to conform locally to a non-Gaussian or multimodal objective landscape. Together with the realism term, this adaptation reduces exploration of weakly supported latent regions.

To reduce computational cost, history matching is performed in two stages. First, each zone-specific generator is optimised independently against the static mismatch and geological-realism terms, yielding latent vectors whose decoded fields approximately honour the well-log data while remaining in well-supported regions of the latent space. Unlike conventional geostatistical conditioning, this static match is approximate because the Decoder does not impose well data as hard constraints. Second, CMA-ES is initialised from the best first-stage solution and jointly minimises the full objective across all generators, including the dynamic mismatch term. Only the second stage requires flow simulation of the assembled reservoir model. Three independent optimisation runs produce three matched realisations.

The OF is written as
\begin{equation}
\mathrm{Loss}
=
w_f \mathrm{Loss}_{flow}
+
w_s \mathrm{Loss}_{static}
+
w_r \mathrm{Loss}_{realism},
\label{eq:objective_function}
\end{equation}
where $\mathrm{Loss}_{flow}$ is the dynamic production mismatch, $\mathrm{Loss}_{static}$ is the static well data mismatch, and $\mathrm{Loss}_{realism}$ is the realism penalty, weighted by $w_f$, $w_s$, and $w_r$ \citep{shishaev2025history}.

The static term is the mean squared error between the decoded property values and the well log values at the cells crossed by wells. This should not be interpreted as strict hard conditioning: a well log has a much smaller physical support volume than a simulation cell, so the exact match between a log measurement and a cell averaged property would be too strong \citep{shishaev2025history}. The proposed formulation allows small deviations and treats them as  small scale uncertainty. The dynamic term is the mean squared error between simulated and observed oil and water production rates, and is evaluated only in the second optimisation stage that implies reservoir simulation.

The realism term penalises latent vectors located in less dense regions towards the edge of the latent space \citep{shishaev2025history}. The \textit{Decoder} induces a local scalar metric in latent space neighbourhood $\mathcal{N}(t)$ through its Jacobian \citep{arvanitidis2018}:
\begin{equation}
\mathrm{Loss}_{static}(\mathbf{z})
=
J_f(\mathbf{z})^T J_f(\mathbf{z}).
\label{eq:latent_metric}
\end{equation}

Ideally, the Jacobian-based metric should increase with distance from the training data, so that the realism penalty is larger in poorly supported regions. Because neural networks extrapolate poorly, the Jacobian term alone does not guarantee this. We therefore add a radial-basis-function precision network \citep{que2016} that estimates the decoder uncertainty as the inverse of a precision centred on the encoded training data. This uncertainty is low near the data and grows with distance, so the realism penalty increases as the latent vector moves away from the training data. The static and dynamic terms guide the optimiser towards models that match the observations, while the realism term discourages geologically implausible solutions.

%% file: chapters/Application.tex
\section{Application}
\label{sec:application}

This section applies the GWAE AHM workflow to the Brugge benchmark, examining latent-space diagnostics, generator quality, and history matching in the hybrid latent space.

\subsection{Latent space analysis and motivation for multiple generators}
\label{sec:schelde_latent}

A geological prior may contain realisations generated using alternative geostatistical algorithms that represent the same depositional setting through different spatial assumptions. In the Schelde diagnostic dataset, OM constructs explicit channel objects, whereas SIS simulates facies cell by cell using variogram-controlled spatial continuity. To test whether both modelling strategies can be represented within one coherent latent space, a single GWAE was trained on a balanced dataset of 2500 OM realisations and 2500 SIS realisations. A latent dimension of 100 was used, larger than that of the final Schelde generator, to reduce the likelihood that any observed separation resulted simply from insufficient latent capacity.

\begin{figure}[htbp]
\centering
\includegraphics[width=1.0\linewidth]{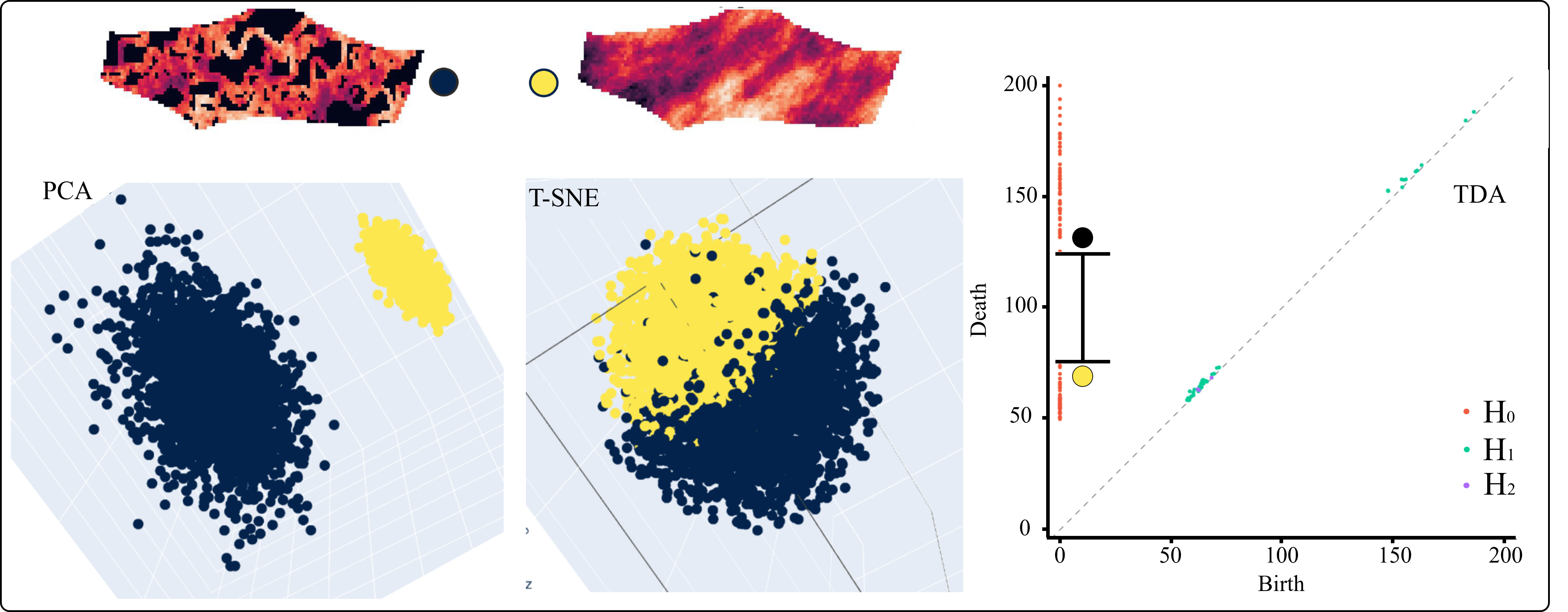}
\caption{Latent space diagnostics for one GWAE trained on the combined Schelde prior. The PCA and t-SNE projections show two separated regions that correspond to the two modelling algorithms: OM realisations shown by blue markers and SIS realisations shown by yellow markers. The latent point cloud is not represented as a single connected region, as evidenced by the two persistent $H_0$ components shown as red dots in the TDA diagram. Adapted from \citet{shishaev2024thesis}.}
\label{fig:schelde_combined_ls}
\end{figure}

Figure~\ref{fig:schelde_combined_ls} summarises the latent-space diagnostics for the combined Schelde dataset. The PCA projection separates the encoded realisations into two distinct clusters corresponding to the two modelling algorithms. The t-SNE projection also shows two compact regions, with the separation becoming apparent when the points are coloured by the modelling algorithm. Topological data analysis, TDA, confirms the same structure: the $H_0$ persistence diagram contains two long-lived connected components, shown as red dots, indicating that the latent point cloud is not represented as a single connected region. Taken together, these observations show that the combined Schelde prior is represented as two distinct latent subpopulations rather than one coherent region.

For history matching, persistently separated components of the latent point cloud can make interpolation between the modelling groups unreliable: optimisation paths may cross regions with little training support, where the \textit{Decoder} has not learned a well-constrained mapping to geological models. Hence, geological realism control in GWAE AHM is imposed on the optimisation by the shape of the latent space with a penalty for moving beyond its edges (see $\mathrm{Loss}_{realism}$ in eq. 5). The shape and the density decay towards the edge of the latent space are defined during GWAE learning.

\begin{figure}[htbp]
\centering
\includegraphics[width=1.0\linewidth]{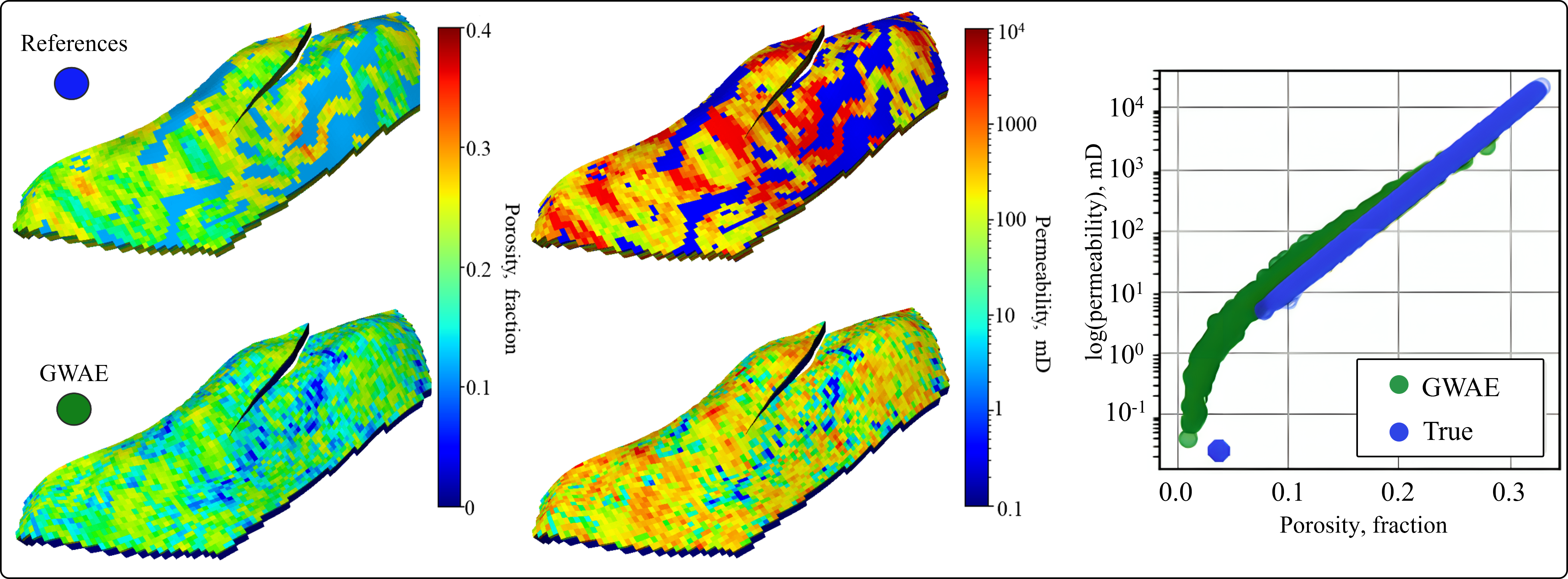}
\caption{Reconstruction failure for one GWAE trained on the combined Schelde prior. Left column: original OM realisation (top) and its GWAE reconstruction (bottom), porosity field. Middle column: the same realisation, permeability field. Right panel: the porosity and permeability relation changes from a two-group structure to a continuous trend. Adapted from \citet{shishaev2024thesis}.}
\label{fig:schelde_combined_quality}
\end{figure}

When an OM Schelde realisation is encoded and reconstructed by the single GWAE, the reconstruction retains the overall property range but substantially smooths the sharp object-defined channel boundaries and the uniform low-property background of the original realisation (Figure~\ref{fig:schelde_combined_quality}). The reconstructed property fields are smoother and visually resemble the SIS realisations more closely. In the porosity-permeability cross-plot, the main trend is retained, whereas the overlapping OM points at a single low-property coordinate (blue) are replaced in the reconstruction by a continuous low-property tail (green). 

The single generator therefore fails on two levels: the latent point cloud splits into separated components, and the \textit{Decoder} cannot reliably reconstruct the channel and shale structure of the OM subpopulation -- a failure detected before any flow simulation. The zone-specific Schelde GWAE (Figure~\ref{fig:pred_true_zones}) retains the channelised pattern more faithfully. The four-zone Brugge formulation follows the prior construction and is assessed through the zone-wise diagnostics below.

\subsection{Generative quality of the GWAE ensemble}
\label{sec:results_generative_quality}

\begin{figure}[htbp]
\centering
\includegraphics[width=1.0\linewidth]{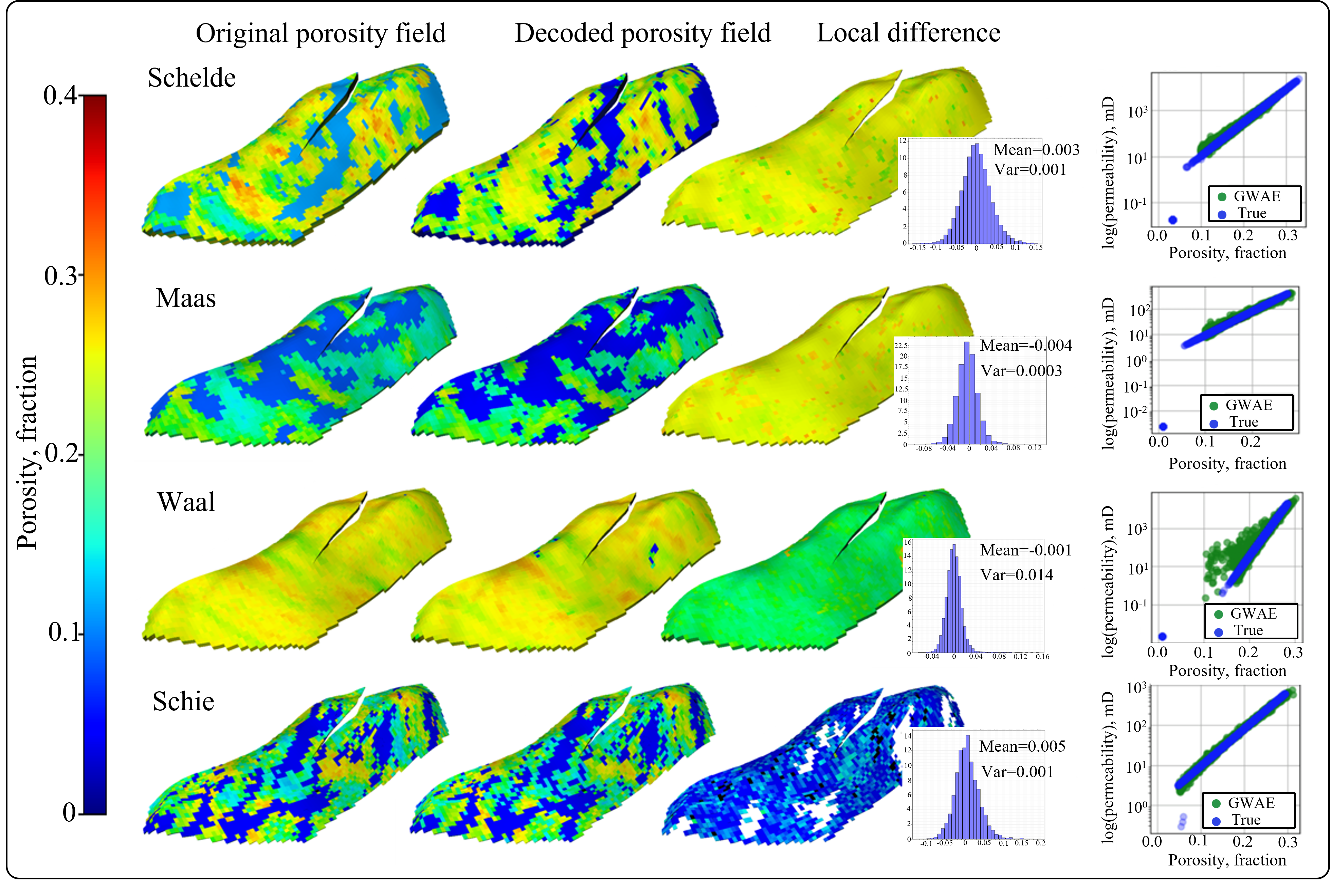}
\caption{Reconstruction quality for the four Brugge zones. Each row corresponds to one stratigraphic zone. From left to right, the columns show the original porosity field, the decoded reconstruction, the local difference, the error histogram with mean and variance, and the porosity-permeability relation for the original and reconstructed samples, respectively. Adapted from \citet{shishaev2024thesis}.}
\label{fig:pred_true_zones}
\end{figure}

The generative quality was evaluated by comparing original prior realisations with their decoded reconstructions. The analysis used porosity fields, reconstruction error, porosity and permeability relations, and latent space diagnostics. Figure~\ref{fig:pred_true_zones} shows that the decoded realisations reproduce the main spatial structure of the original models in all four stratigraphic zones. The reconstruction error is centred close to zero, which indicates that there is no strong systematic bias in the recovered porosity fields. The error is distributed evenly across the porosity field rather than localised in specific regions, so the remaining differences do not form large coherent artefacts.

The porosity--permeability cross-plots in Figure~\ref{fig:pred_true_zones} show that the zone GWAEs retain the main petrophysical trends of the prior without any explicit porosity--permeability constraint in the training loss. Reconstructed samples remain close to the original distributions for Schelde, Maas, and Schie, whereas for Waal the reconstruction preserves the central trend but shows a broader spread, particularly at lower permeability and intermediate porosity.

\begin{figure}[htbp]
\centering
\includegraphics[width=1.0\linewidth]{"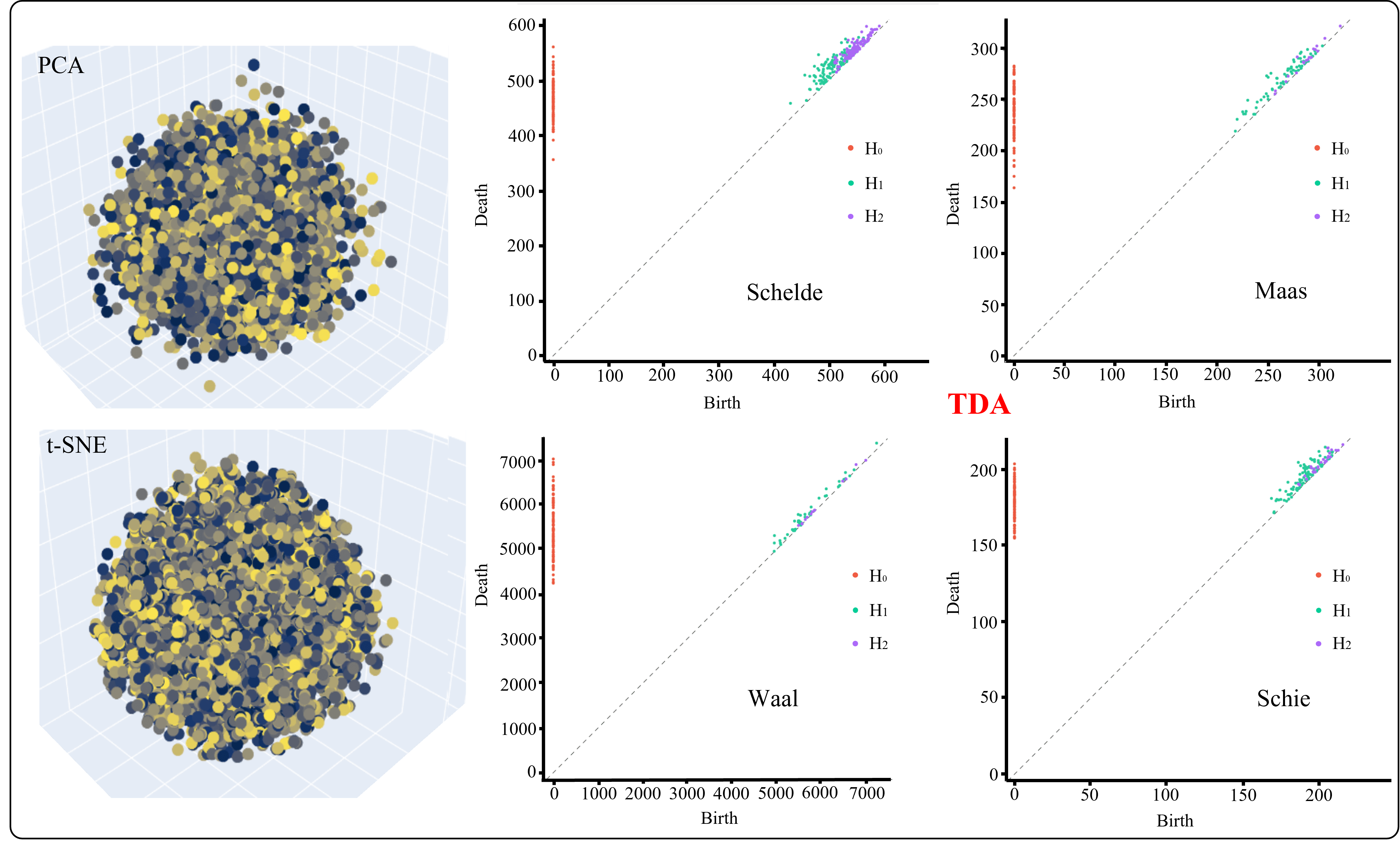"}
\caption{Latent space diagnostics for the four Brugge zone generators. The left panels show PCA (top) and t-SNE (bottom) projections for the Schelde zone, coloured by the major range of the porosity variogram. The right panels show TDA diagrams for Schelde, Maas, Waal, and Schie. Each zone is represented by one dominant connected region, which supports the use of one generator per zone. Adapted from \citet{shishaev2024thesis}.}
\label{fig:brugge_ls}
\end{figure}

Figure~\ref{fig:brugge_ls} shows the latent spaces of the four individual generators. In the PCA and t-SNE projections of Schelde, points coloured by the major range of the porosity variogram are intermixed with no clear gradient; the same holds for the other sampled parameters (Table~\ref{tab:prior_ranges}). Unlike the combined Schelde prior (Section~\ref{sec:schelde_latent}), each zone has one dominant connected region, supporting optimisation without crossing gaps between separated prior groups. The corresponding PCA and t-SNE projections for Maas, Waal, and Schie are provided in Figures~S1--S3 of the Supplementary Material.

The persistence diagrams are used here to assess the connectedness of each zone-wise latent point cloud. In all four zones, the $H_0$ features merge into one dominant connected component, and no second long-lived component comparable to that of the mixed OM/SIS Schelde prior is observed. The raw filtration radii differ between zones because the four GWAEs were trained independently and use different latent dimensions and numerical coordinate scales. These radii represent distances in latent space and should not be interpreted as geological length scales or as direct measures of geological heterogeneity. TDA therefore supports the absence of persistent separation within each zone-wise prior, but it does not by itself identify the depositional or spatial geological characteristics learned by the generator.

\subsection{History matching in the hybrid latent space}
\label{sec:results_history_matching}

History matching was performed in the 160-dimensional hybrid latent space following the procedure defined in Section~\ref{sec:optimisation}. A reference Brugge realisation was treated as unknown, except for static well data and production response.

\begin{figure}[htbp]
\centering
\includegraphics[width=1.0\linewidth]{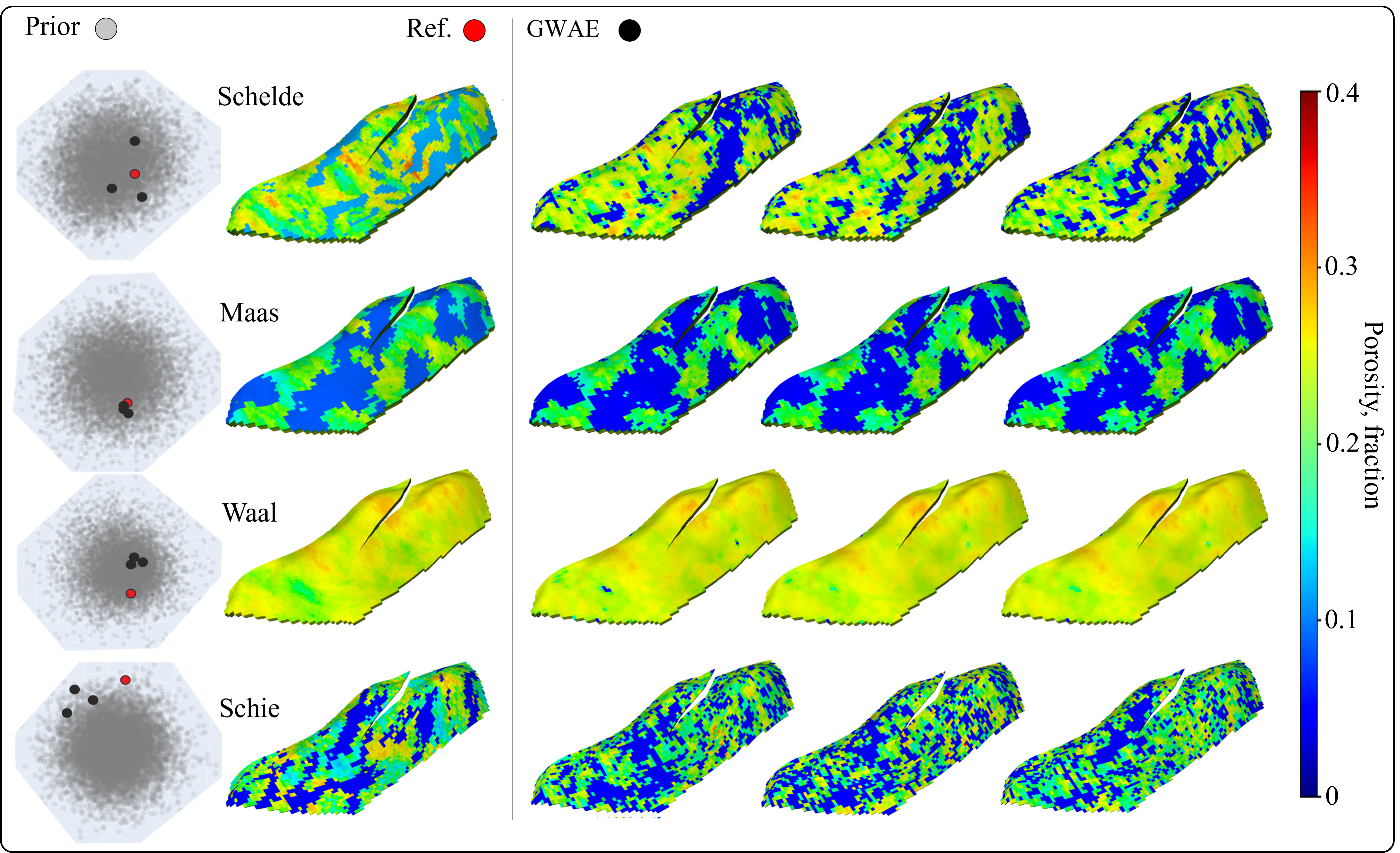}
\caption{History matching results in the hybrid latent space. Each row shows one stratigraphic zone. Left column: the prior cloud in the PCA projection (grey), the reference latent position (red), and the three CMA-ES solutions (black). Middle column: reference porosity field. Right three columns: GWAE-decoded porosity fields for the three matched solutions. Adapted from \citet{shishaev2024thesis}.}
\label{fig:hm_zones}
\end{figure}

Figure~\ref{fig:hm_zones} shows the matched solutions in the latent projections and in the property domain. In all four zones, the optimised latent codes remain within the projected support of the prior ensemble, but their proximity to the encoded reference differs between zones. The Maas solutions lie close to the reference in the PCA projection, whereas the Waal solutions form a compact group at a visibly different projected location. Nevertheless, the Waal reconstructions retain the main large-scale property trend and give a small pore-volume error. This illustrates the non-uniqueness of the inverse problem and shows that distance in a low-dimensional PCA projection should not be treated as a direct measure of geological or dynamic similarity. At the same time, the location of the optimised codes within the projected prior clouds is consistent with the intended role of the realism term: it discourages the search from moving towards poorly supported regions of the learned latent space without preventing improvement of the static and dynamic data match.

The recovered porosity fields shown in Figure~\ref{fig:hm_zones}  display two main behaviour patterns. The matched Maas and Waal realisations reproduce the main large-scale porosity patterns of the reference. Waal is recovered particularly well, which is important because it contributes the dominant share of total fluid production, and the matched Waal models preserve the smooth lateral property trend of the reference model. Schelde is more difficult to recover because its fluvial architecture contains abrupt property contrasts and uncertain channel connectivity. The well data provide only sparse local conditioning, so channel placement and the associated permeability connectivity between wells remain weakly constrained. Consequently, matched Schelde realisations can remain consistent with the available static and dynamic data while differing from the reference away from the wells.

For Schie, the encoded reference lies near the edge of the projected prior cloud. Because Schie is thin and contributes less than 1\% of total fluid production, both the static and dynamic terms carry little information about this zone, leaving the optimisation weakly constrained. With almost no data-driven signal to guide the search, the recovered Schie fields are noisier than the reference.

%% file: chapters/Results.tex
\section{Results}
\label{sec:results}

This section assesses the history-matching results in terms of data agreement and consistency with the geological prior. The analysis covers optimisation behaviour, bulk-property statistics, production response, and a qualitative comparison with published Brugge HM studies.

\begin{figure}[htbp]
\centering
\includegraphics[width=\linewidth]{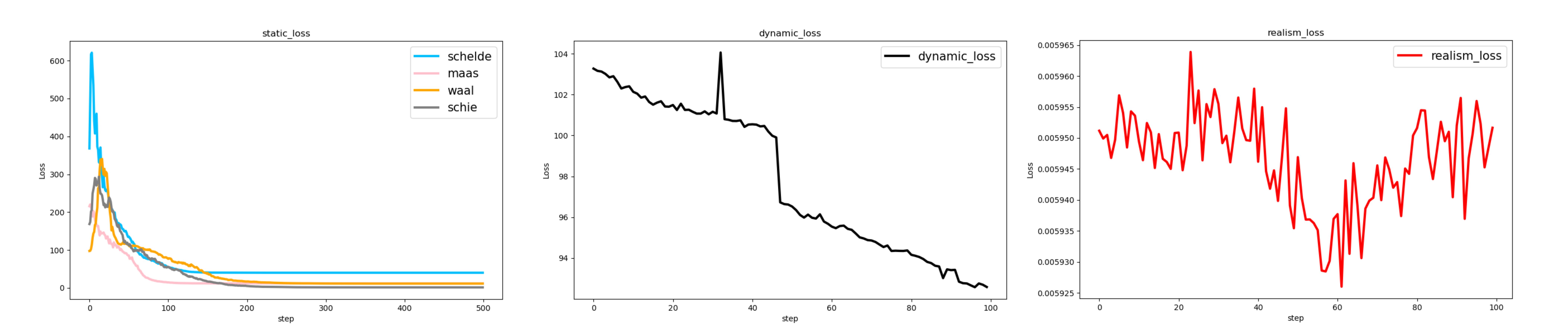}
\caption{Objective-function histories during two-stage history matching. Left: static mismatch for the four zones during the first 500 iterations. Centre and right: dynamic mismatch and the realism penalty during the following 100 interations.  Adapted from \citet{shishaev2024thesis}.}
\label{fig:hm_loss_components}
\end{figure}

Figure~\ref{fig:hm_loss_components} shows the loss function evolution through CMA-ES optimisation: static mismatch drops sharlpy and settles across the four startigraphy zone (in the fluvial section the wells are matched the relatively poorer due to higher heterogeneity). Once static loss is settled within the initial 400 iterations, the dynamic misfit takes 100 additional interations to reach the acceptable level of match. The realism loss term acts as a regulariser -- it controls the realism within the latern space bounds rather than being minimised.

\begin{figure}[htbp]
\centering
\includegraphics[width=1.0\linewidth]{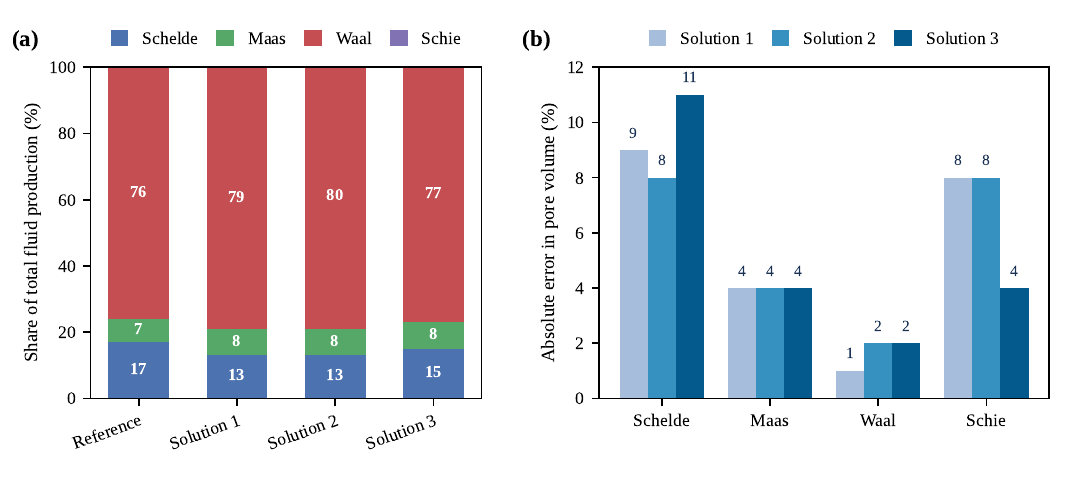}
\caption{Reservoir statistics of the matched ensemble against the reference model, from the values in Table S1. (a) Share of total fluid production by stratigraphic zone for the reference model and the three CMA-ES solutions; Waal dominates in all cases (76 to 80$\%$), and the Schie contribution stays below 1$\%$. (b) Absolute error in pore volume relative to the reference, by zone, for the three solutions.}
\label{fig:hm_stats}
\end{figure}

The mean porosity, porosity standard deviation, and pore-volume error are reported as complementary bulk-property diagnostics rather than as measures of dynamic match. The matched models reproduce the reference mean porosity reasonably well, while the porosity variability and pore-volume errors show zone-dependent departures. The absolute pore-volume error is up to 11$\%$ for Schelde, about 4$\%$ for Maas, 1--2$\%$ for Waal, and 4--8$\%$ for Schie (Figure~\ref{fig:hm_stats}(b)). These results indicate that the matched solutions remain broadly consistent with the reference in terms of bulk porosity and pore volume. Dynamic performance is assessed separately from the oil and water responses, which depend jointly on pore volume, permeability, connectivity, and phase-flow parameters.

\begin{figure}[htbp]
\centering
\includegraphics[width=1.0\linewidth]{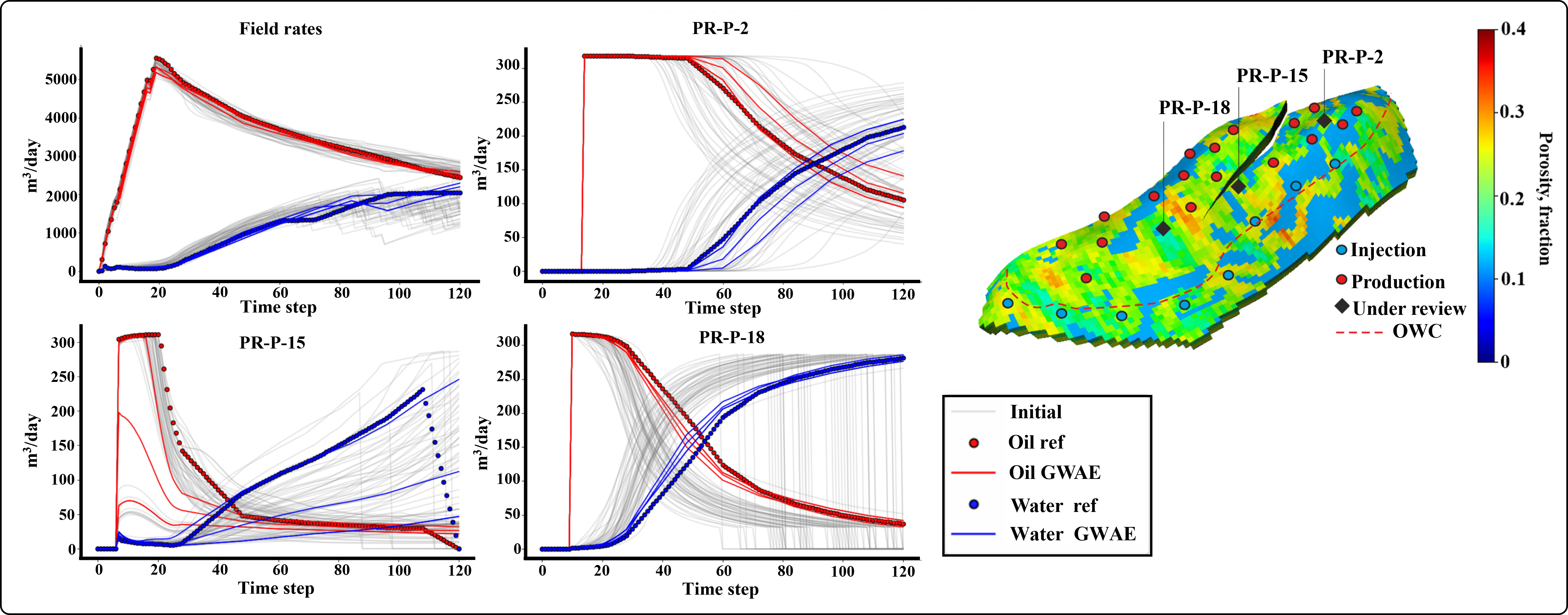}
\caption{Dynamic match for the Brugge case. The panels show field-level oil and water rates and the well-level responses of PR-P-2, PR-P-15, and PR-P-18. Grey curves represent the prior ensemble, coloured curves represent the three matched GWAE solutions, and points represent the reference response. The right panel shows the locations of the selected wells on the reference porosity field. Adapted from \citet{shishaev2024thesis}.}
\label{fig:hm_rates}
\end{figure}

Figure~\ref{fig:hm_rates} shows that prior coverage and matching quality differ between the field and well scales. At the field level, the reference oil and water responses are largely contained within the broad prior envelope, and the three matched solutions concentrate close to the observed profiles. At the well level, the coverage is less uniform because individual responses depend on local permeability and connectivity. In particular, the PR-P-18 history is not fully covered by the sampled prior ensemble, indicating a local bias in the prior dynamic response and making this well more difficult to match. The optimised responses nevertheless move towards the reference trend, although a larger residual mismatch remains where the history lies near or outside the prior envelope. Part of the remaining oil-water mismatch is also expected because only porosity and permeability are updated, while relative permeability and other phase-flow parameters remain fixed.

\begin{figure}[htbp]
\centering
\includegraphics[width=1.0\linewidth]{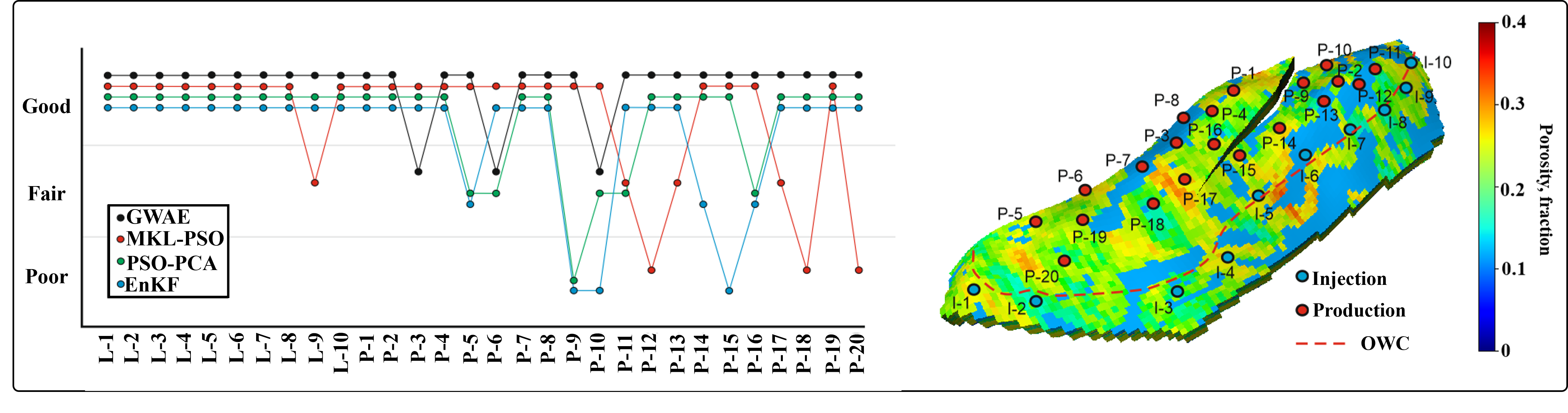}
\caption{Qualitative comparison of history matching quality for GWAE, MKL-PSO, PSO-PCA, and EnKF on the Brugge case. The comparison uses the well-level good, fair, and poor match classification reported in previous Brugge studies and should be interpreted as a qualitative visual comparison rather than a controlled quantitative benchmark. The right panel shows the map of production and injection well locations on the reference porosity field. Adapted from \citet{shishaev2024thesis}.}
\label{fig:comparison_methods}
\end{figure}

Figure~\ref{fig:comparison_methods} places the GWAE results in the context of previously published Brugge history matching studies that report a well level good, fair, and poor classification: EnKF, Particle Swarm Optimisation with PCA parameterisation (PSO-PCA), and Multiple Kernel Learning with Particle Swarm Optimisation (MKL-PSO) \citep{peters2010results, mohamed2010application, demyanov2015mkl}. More recent Brugge studies that combine production data with tracer or 4D seismic data are not included in the comparison because they do not report the same well-level classification \citep{cruz2022joint, Chalub_Brugge}. The published studies differ in their calibration variables, objective functions, data types, optimisation strategies, and reporting metrics, so the figure should be interpreted as a qualitative context rather than as a controlled benchmark.

Within this limitation, as shown in Figure~\ref{fig:comparison_methods}, the GWAE workflow achieves a competitive well-level match, with most wells falling into the good match category, and the number of poor matches is limited compared with the three reference methods considered. The more pivotal difference, however, is methodological. EnKF, PSO-PCA, and MKL-PSO update reservoir model parameters through different reduced or ensemble formulations, but they do not explicitly constrain every accepted model to remain inside a learned geological prior manifold. In the proposed GWAE workflow, the optimiser searches in the latent space of trained graph generative models, and the realism term penalises poorly supported latent regions. The comparison, therefore, indicates not only that GWAE can match production data, but that this match is obtained while maintaining an explicit link to the geological prior and while producing an ensemble of plausible matched realisations.

These results show that the proposed workflow is effective for the Brugge benchmark. The graph generators operate on the native irregular grid, the multiple generator formulation gives a coherent hybrid latent space, and the optimiser finds matched models that reproduce both static and dynamic data while remaining connected to the geological prior. The remaining errors concentrate in zones with strong local heterogeneity (Schelde) or weak contribution to production (Schie), which suggests that further improvement is more likely to come from extending the dataset or the calibration variables than from changing the latent parameterisation itself.

%% file: chapters/Conclusions.tex
\section{Conclusion}
\label{sec:conclusion}

This study addressed assisted history matching of field-scale reservoir models with a heterogeneous, multi-zone geological prior on an irregular corner-point grid. History matching was formulated as a search in a hybrid latent space built from zone-specific graph-based Wasserstein autoencoders (GWAEs) operating directly on the native Brugge grid. The four zone latent spaces were combined into a 160-dimensional joint space and optimised with CMA-ES against an objective integrating static mismatch, dynamic mismatch, and geological realism; the need for this multiple-generator decomposition was itself established using latent-space diagnostics, including persistent homology.

The main findings are as follows:
\begin{itemize}
    \item \textbf{One generator is not sufficient for a heterogeneous prior.} For the combined Schelde prior, which includes realisations generated by object modelling and sequential indicator simulation, a single GWAE produced a latent point cloud with two persistent $H_0$ components. The \textit{Decoder} also failed to preserve the channel and shale structure of the object modelling realisations.
    
    \item \textbf{A multiple-generator decomposition improves latent coherence.} The four zone GWAEs each produced one dominant connected latent component. They reconstructed the main spatial patterns of their priors and preserved the porosity and permeability relation implicitly, without adding this relation as an explicit constraint in the loss function.
    
    \item \textbf{The hybrid latent space supports history matching while remaining within the prior manifold.} The matched solutions remained within the projected prior clouds in all four zones and reproduced field-level oil and water rates within the prior spread. Absolute pore-volume error stayed within roughly 1--11\% across zones, lowest for Waal and highest for Schelde.
    
    \item \textbf{The workflow provides a competitive well-level match while keeping an explicit link to the geological prior.} Most wells fall into the good match category according to the qualitative classification used in earlier Brugge studies. Unlike Ensemble Kalman Filter (EnKF), Particle Swarm Optimisation with principal component analysis of parameterisation (PSO-PCA), and Multiple Kernel Learning with Particle Swarm Optimisation (MKL-PSO), every accepted model in the proposed workflow remains constrained by the learned generative prior manifold.
\end{itemize}

The current formulation assumes independent geological priors across zones; reservoirs with cross-zone dependencies would require shared latent variables or conditional generators. Calibration updates only porosity and permeability, leaving relative permeability and other phase-flow parameters fixed. Extending this variable set may improve flexibility, although the present results do not isolate the causes of residual well-level mismatch. Static conditioning is approximate, and its mean-squared-error formulation does not explicitly model the support difference between well logs and grid-cell properties.

The study considers a limited space of uncertainty associated with primarily 2-point spatial correlation parameters used to design the prior ensemble of models. However, the framework demonstrated its generic capability to handle more diverse prior definitions and has the topological diagnostics to differentiate between multiple geological concepts. 
 

Further work will enhance the method with posterior sampling in the latent space for a more accurate evaluation of the prediction credible intervals, not just several exemplary solutions.  Additional observed data, such as 4D seismic or tracer data, can be also integrated into the framework.

%% file: chapters/Acknowledgments.tex
\section{Acknowledgements}
The authors appreciate the support of Rock Flow Dynamics for GeoDataScience \& UQ group research at Heriot-Watt University and the provision of tNavigator software licences for reservoir flow simulations.
The authors are grateful to Gleb Shishaev's PhD examiners, Prof. Ahmed ElSheikh of Heriot-Watt University and Prof. Denis Voskov of TU Delft for their careful review of the thesis and for providing valuable comments and advice that helped improve this paper.

%% file: sup_materials.tex
\clearpage
\section{Supplementary Materials}

\setcounter{table}{0}
\setcounter{figure}{0}
\renewcommand{\thetable}{S\arabic{table}}
\renewcommand{\thefigure}{S\arabic{figure}}
\ifdefined\theHtable
  \renewcommand{\theHtable}{supp.\arabic{table}}
\fi
\ifdefined\theHfigure
  \renewcommand{\theHfigure}{supp.\arabic{figure}}
\fi

\begin{table}[htbp]
\caption{Main characteristics of the porosity property for the four
stratigraphic zones. For each zone the reference model and three GWAE
optimised realisations are reported: average porosity (sand only), porosity
standard deviation, pore volume, absolute error of pore volume relative to the
reference model, and share of total fluid production.}
\label{tab:poro_char}
\begin{center}
\begin{adjustbox}{max width=\textwidth}
\begin{tabular}{ | c | c | c | c | c | c | }
\hline
\textbf{Realisation}
 & \shortstack{\textbf{Av.\ Poro.}\textsuperscript{*}\\\textbf{(frac)}}
 & \shortstack{\textbf{Std.}\\\textbf{Poro.}}
 & \shortstack{\textbf{Pore Vol.}\\($10^{6}\,\mathrm{m}^{3}$)}
 & \shortstack{\textbf{Abs.\ Err.}\\\textbf{(\%)}}
 & \shortstack{\textbf{\% of Total}\\\textbf{Fluid Prod.}} \\
\hline
\multicolumn{6}{| c |}{\textbf{Schelde}} \\
\hline
Reference       & 0.21 & 0.04 & 53  & --  & 17 \\
\hline
GWAE optimised  & 0.20 & 0.04 & 48  & 9   & 13 \\
\hline
GWAE optimised  & 0.20 & 0.04 & 49  & 8   & 13 \\
\hline
GWAE optimised  & 0.20 & 0.04 & 47  & 11  & 15 \\
\hline
\multicolumn{6}{| c |}{\textbf{Maas}} \\
\hline
Reference       & 0.16 & 0.04 & 80  & --  & 7 \\
\hline
GWAE optimised  & 0.16 & 0.03 & 77  & 4   & 8 \\
\hline
GWAE optimised  & 0.16 & 0.03 & 77  & 4   & 8 \\
\hline
GWAE optimised  & 0.16 & 0.03 & 77  & 4   & 8 \\
\hline
\multicolumn{6}{| c |}{\textbf{Waal}} \\
\hline
Reference       & 0.23 & 0.03 & 195 & --  & 76 \\
\hline
GWAE optimised  & 0.24 & 0.02 & 197 & 1   & 79 \\
\hline
GWAE optimised  & 0.24 & 0.02 & 198 & 2   & 80 \\
\hline
GWAE optimised  & 0.24 & 0.02 & 198 & 2   & 77 \\
\hline
\multicolumn{6}{| c |}{\textbf{Schie}} \\
\hline
Reference       & 0.18 & 0.06 & 25  & --  & $<1$ \\
\hline
GWAE optimised  & 0.18 & 0.04 & 23  & 8   & $<1$ \\
\hline
GWAE optimised  & 0.19 & 0.05 & 27  & 8   & $<1$ \\
\hline
GWAE optimised  & 0.18 & 0.04 & 24  & 4   & $<1$ \\
\hline
\end{tabular}
\end{adjustbox}
\end{center}

\vspace{4pt}
{\footnotesize\textsuperscript{*}Average porosity values are for sand only.}
\end{table}


\begin{figure}
    \centering
    \includegraphics[width=1\textwidth]{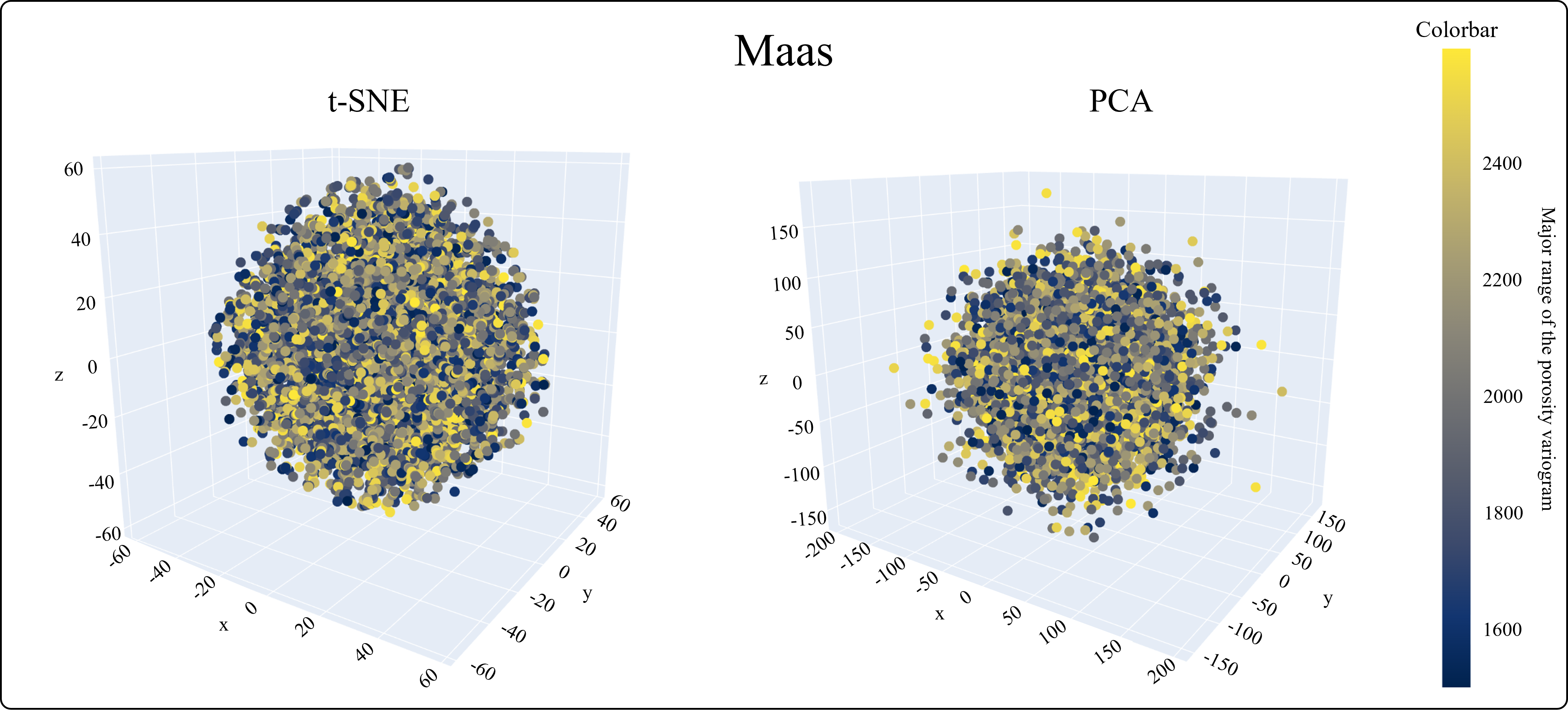}
    \caption{Latent space of the Maas zone generator: t-SNE (left) and PCA
    (right) projections of the encoded prior realisations, coloured by the major
    range of the porosity variogram. As for the Schelde zone in the main text,
    the points are intermixed with no clear gradient, so the latent cloud is not
    organized by this parameter.}
    \label{fig:ls_maas}
\end{figure}


\begin{figure}
    \centering
    \includegraphics[width=1\textwidth]{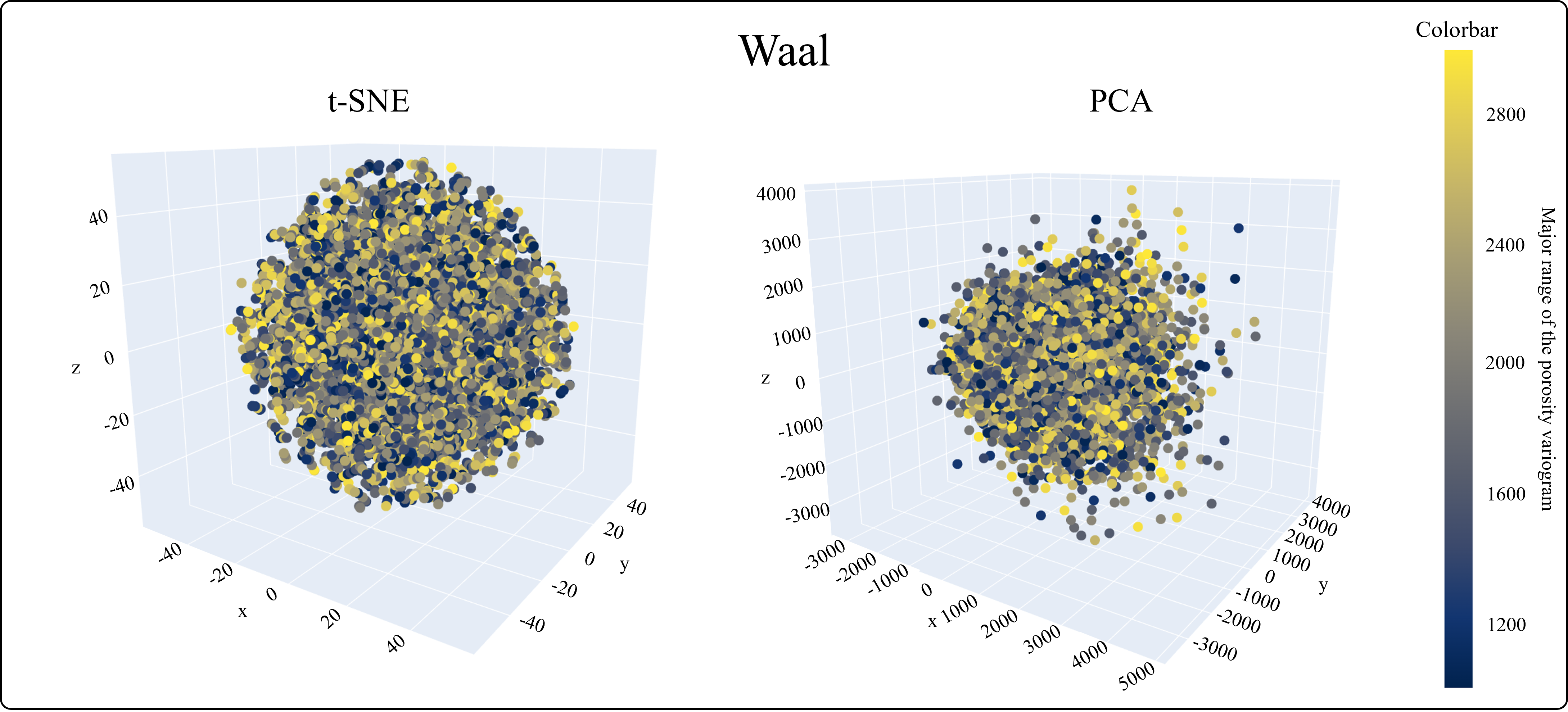}
    \caption{Latent space of the Waal zone generator: t-SNE (left) and PCA
    (right) projections of the encoded prior realisations, coloured by the major
    range of the porosity variogram. The points are intermixed with no clear
    gradient, so the latent cloud is not organized by this parameter.}
    \label{fig:ls_waal}
\end{figure}


\begin{figure}
    \centering
    \includegraphics[width=1\textwidth]{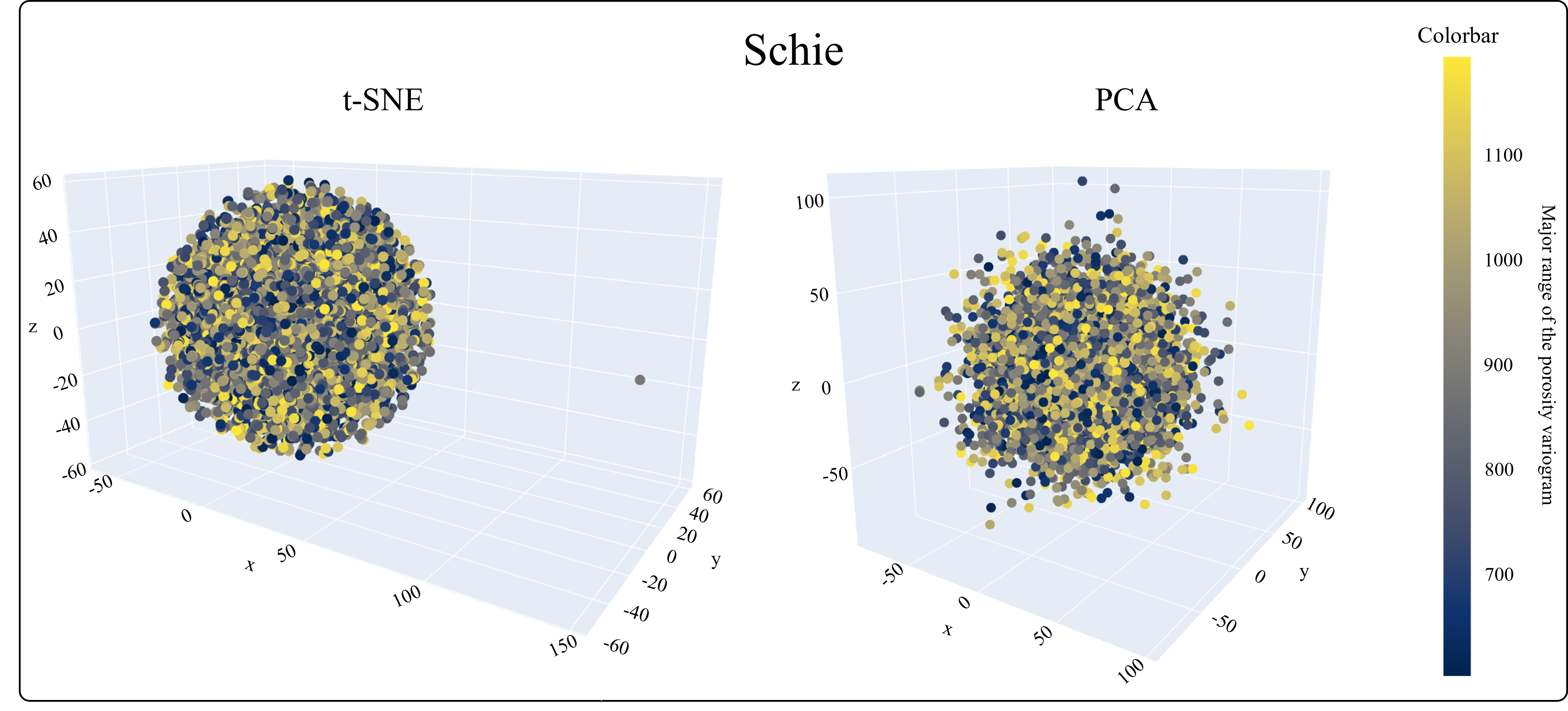}
    \caption{Latent space of the Schie zone generator: t-SNE (left) and PCA
    (right) projections of the encoded prior realisations, coloured by the major
    range of the porosity variogram. The points are intermixed with no clear
    gradient, so the latent cloud is not organized by this parameter.}
    \label{fig:ls_schie}
\end{figure}